\documentclass[11pt]{elsarticle}
\usepackage[utf8]{inputenc}
\usepackage[english]{babel}
\usepackage{enumerate}
\usepackage{color}
\usepackage{graphicx}
\usepackage{amssymb, amsmath, amsthm}
\usepackage{amsfonts}
\usepackage{bm}
\usepackage{calc}
\usepackage{caption}
\usepackage{subcaption}
\usepackage{url}
\usepackage{epstopdf}
\usepackage{float}

\usepackage{geometry}
\usepackage[title]{appendix}
\usepackage{mathtools}

\numberwithin{equation}{section}

{\begin{color}{red}\par\noindent\textbf{Kommentar}\begin{itshape}\par\noindent}
{\end{itshape}\end{color}}

\theoremstyle{remark}
\newtheorem{remark}{Remark}[section]

\theoremstyle{definition}

\theoremstyle{definition}
\newtheorem*{definition*}{Definition}

\begin{document}
\title{Data-driven discovery of PDEs in complex datasets}
\date{}

\author{Jens Berg}
\ead{jens.berg@math.uu.se}
\author{Kaj Nystr\"{o}m}
\ead{kaj.nystrom@math.uu.se}
\address{Department of Mathematics, Uppsala University\\
SE-751 05 Uppsala, Sweden}

\begin{abstract}
Many processes in science and engineering can be described by partial differential equations (PDEs). Traditionally, PDEs are derived by considering first principles of physics to derive the relations between the involved physical quantities of interest. A different approach is to measure the quantities of interest and use deep learning to reverse engineer the PDEs which are describing the physical process.

In this paper we use machine learning, and deep learning in particular, to discover PDEs hidden in complex data sets from measurement data. We include examples of data from a known model problem, and real data from weather station measurements. We show how necessary transformations of the input data amounts to coordinate transformations in the discovered PDE, and we elaborate on feature and model selection. It is shown that the dynamics of a non-linear, second order PDE can be accurately described by an ordinary differential equation which is automatically discovered by our deep learning algorithm. Even more interestingly, we show that similar results apply in the context of more complex simulations of the Swedish temperature distribution.
\end{abstract}

\begin{keyword}
Machine learning \sep Deep learning \sep Partial differential equations \sep Neural networks
\end{keyword}

\maketitle

\section{Introduction} \label{introduction}
Modern technology has made high-quality data available in abundance. It is estimated that more than 2.5 quintillion bytes of data is generated every day and that 90\% of all data were generated in the last two years alone \cite{ibmdata}. The amount of user generated data on social media and data generated through smart sensors in the Internet of things will likely contribute to an even faster increase. A major problem of scientific and industrial interest is how to transform the data into a predictive model which can give insights on the data generating process.

The data generating process in the natural sciences is often described in terms of differential equations. There is a vast amount of literature spanning over decades available for the identification of dynamical systems where the quantities of interest are measured as a function of time or some other dependent variable. See for example \cite{identificationsurvey, identificationstoica, identificationmodeling, motions, identificationbook, identificationblackbox, identificationblackboxmath}. The identification of time-dependent partial differential equations (PDEs) through data analysis is an emerging and exciting field of research which is not as explored as dynamical systems. The research has been made available through the recent progress in machine learning algorithms and their efficient implementation in open source software.

PDEs are traditionally derived by considering first physical principles. For example the Navier-Stokes equations in fluid dynamics are derived by considering the conservation of mass, momentum, and energy for a control volume in a fluid. There are, however, many situations where derivations by first principles are intractable or even impossible as they become too complicated or the governing physical laws are unknown. In such situations there are typically several geostationary points where changes of a quantities of interest are measured over time. Datasets consisting of such spatio-temporal data is the interest of this paper and we aim to develop methods which can automatically identify a PDE which is generating the dataset.

The emerging field of data-driven discovery of PDEs can be split into three approaches: (1) Sparse regression, (2) Gaussian processes, and (3) Artificial neural networks. Sparse regression is based on a library of candidate terms and sparse model selection to select the most important terms \cite{sparsepde, datadriven1, haydensparsepde}. Identification using Gaussian processes works by placing a Gaussian process prior on the unknown coefficients of the PDE and infer them by using maximum likelihood estimation \cite{gausspde1, gausspde2, gausspde3}. Artificial neural networks can be used as sparse regression models, act as priors on unknown coefficients, or completely determine a general differential operator \cite{inverse, neuraldiscovery1, neuraldiscovery2}.

In this paper we will focus on deep neural networks to extend and complement previous work mentioned in the above references. There are two main contribution in this paper. The first is that we use a unified neural network approach for both sparse regression and the identification of general differential operators. The second is that we include complex datasets where necessary transformations of the input data manifest as coordinate transformations which yield metric coefficients in the identified PDE.

\section{Method} \label{method}
We are working under the assumption that we have an unordered dataset consisting of space-time coordinates and function values where the governing equation is unknown. The goal is to identify a PDE which approximately has the function values as the solution in the space-time points. The first step is to fit a function to the data which can be used to compute the derivatives with respect to the space-time coordinates. This is a separate preprocessing step and any method can be used, for example finite differences \cite{pdeidentificationfinitedifference}, polynomial interpolation \cite{pdeidentificationpolynomial}, finite elements, spectral methods, radial basis functions, or neural networks. The most recent work have been focused on polynomial interpolation \cite{datadriven1} or neural networks \cite{2017arXiv171110561R, neuraldiscovery2} due to their independence of structured data and insensitivity to noise. The identified PDE depends highly on the quality of the approximating function and a comparative study of various approximation methods would be valuable and is the topic of future research. We will use deep neural networks as approximating functions. Deep neural networks are universal smooth function approximators \cite{Hornik1991, Hornik1990551, Cybenko1989} and their derivatives are analytically available through backpropagation \cite{backprop, efficientbackprop} or automatic differentiation \cite{identificationsurvey} in open source software such as TensorFlow \cite{tensorflow} or PyTorch \cite{pytorch}.

We assume that our data consists of the triplets $t$, $\mathbf{x} = [x_1, x_2, \ldots, x_N]$, and $\mathbf{u} = [u_1, u_2, \ldots, u_M]$ which is describing a vector valued mapping $\mathbf{u}: \mathbb{R}^{N+1} \to \mathbb{R}^M$, where $t$ denotes the time variable, $x_1, \ldots, x_N$ the space variables, and $u_1, \ldots u_M$ the function values. In the first step we approximate the function $u$ by a deep neural network $\hat{u} = \hat{u}(\mathbf{x}, t; \mathbf{p})$ where $\mathbf{p}$ denotes the vector of coefficients in the network. We will usually drop explicit parameter dependence, unless required, to ease the notation. We will use the hyperbolic tangent as activation function and solve the regularized minimization problem for the coefficients,
\begin{equation}
\mathbf{p}^* = \min_{\mathbf{p}} \frac{1}{2}||u(x, t) - \hat{u}(\mathbf{x}, t; \mathbf{p})||^2 + \frac{\alpha_p}{2}||\mathbf{p}||^2,
\label{minu}
\end{equation}
by using the BFGS \cite{bfgs} or L-BFGS \cite{lbfgs} methods for small and large scale problems, respectively. When solving the minimization problem \eqref{minu}, we do not distinguish between the time and space coordinates. Different datasets require different neural networks designs and it would be interesting to try neural networks which are tailored for time-series prediction, for example recurrent neural networks, in this context. Such a study is, however, beyond the scope of this paper.

In the second step we seek a parameterized function $\hat{L} = \hat{L}(\hat{u}, \partial \hat{u}, \ldots, \partial^m \hat{u}; \mathbf{q})$, where the notation $\partial^j \hat{u}$ means all partial derivatives of $\hat{u}$ with respect to $x_1, \ldots, x_n$ up to order $m$ such that
\begin{equation}
\hat{u}_t = \hat{L}(\hat{u}, \partial \hat{u}, \ldots, \partial^m \hat{u}).
\end{equation}
$\hat{L}$ is then the approximation of the, yet unknown, differential operator in the governing PDE. The restriction to first order time derivatives is without loss of generality as we can compute derivatives of any order from the neural network approximation $\hat{u}$.

Depending on the choice of parametrization of $\hat{L}$ it is possible to discover a wide range of PDEs and encompass the methods described in \cite{datadriven1, datadriven2, haydensparsepde, sparsepde, neuraldiscovery1, neuraldiscovery2} in a single framework. The framework we have chosen here is to represent $\hat{L}$ by a feedforward neural network and to find $\hat{L}$ by gradient based optimization. We recover the sparse regression method by having a neural network without hidden layers with candidate terms as input features, in which case the neural network reduces to a linear model. We recover classical PDEs, which are polynomial in $\hat{u}$ and its partial derivatives, by computing all partial derivatives up to some order $m$, all non-linear combinations up to some order $k$, and having them as input features to a linear model. There are
\begin{equation}
\mathcal{M} = M\left(1 +  \sum_{i=1}^m \binom{i + N - 1}{N - 1}\right)
\end{equation}
partial derivative terms up to order $m$ and
\begin{equation}
\mathcal{K} = \sum_{i=1}^k \binom{i + \mathcal{M} - 1}{\mathcal{M} - 1}
\end{equation}
non-linear combinations up to order $k$. For example, the time-dependent compressible Navier-Stokes equations in 3D have $N=3$ space variables, $M=5$ unknowns, non-linear terms up to order $k=2$, and partial derivatives to up order $m=2$. This gives a total of $\mathcal{M} = 50$ partial derivative terms and $\mathcal{K} = 1325$ possible input features. While the number of input features grows combinatorially with the number of partial derivatives and non-linear order, modern day machine learning with neural networks casually deal with input features in the order of million or even billions. Even the most basic standard example of hand written digit recognition using the MNIST dataset has $28 \times 28 = 784$ input features -- the number of pixels of each image in the dataset. Finally, we can let $\hat{L}$ be given by a neural network of arbitrary complexity with the $\mathcal{M}$ partial derivative terms as input features to get an arbitrarily complex differential operator.

There is always a trade-off between model complexity and interpretability. A linear model with candidate terms as input features provides a simple model which can be read, analyzed, and understood. It does, however, require some physical understanding of the data generating process to ensure that the set of input features is sufficient. A general neural network model is on the other extreme. It can approximate an arbitrary complex differential operator but the resulting operator can neither be read nor understood. A linear model with polynomial input features is somewhere in between. Sparse regression with L1 regularization will remove some insignificant terms but some manual post cleaning will probably be required to get a interpretable model. In all cases, the model is unlikely to produce a well-posed PDE in the sense of Hadamard \cite{hadamard}.

As the true differential operator $L$ is not known and we have no training data for it, the goal is to find a set of parameters $\mathbf{q}^*$ such that the residual of the approximate PDE is minimized,
\begin{equation}
\mathbf{q}^* = \min_\mathbf{q} \frac{1}{2}||\hat{u}_t - \hat{L}(\hat{u}, \partial \hat{u}, \ldots, \partial^m \hat{u}; \mathbf{q})||^2 + \frac{\alpha_q}{2}||\mathbf{q}||_1^2.
\label{minpde}
\end{equation}
We typically add regularization in the $L^1$-norm to favor sparsity in the resulting PDE model. The optimization problems \eqref{minu} and \eqref{minpde} are very different from an optimization perspective. The former is a highly non-convex optimization problem over a large number of parameters and a limited amount of data. The latter is, in the linear model case, a convex optimization problem over a small number of parameters and a large amount of data. In the 3D Navier-Stokes example above, let us assume that we have sampled the solution 100 times on a  $32 \times 32 \times 32$ grid. This gives us a dataset of size $3276800 \times 4$ in the optimization of \eqref{minu} and $3276800 \times 1325$ in the optimization of \eqref{minpde}. Data driven discovery of PDEs is thus suitable on heterogeneous systems where the optimization of \eqref{minu} is performed on GPUs with many cores and limited memory while the optimization of \eqref{minpde} is performed on CPUs with few cores and large memory.

\subsection{Feature scaling} \label{scaling}
It is well-known that machine learning algorithms perform poorly unless the input features are scaled correctly. In the previous work on data-driven discovery of PDEs, all data were generated by known PDEs on simple geometries which did not require any transformation of the input features. In real life applications, however, the domain of interest is in general neither simple nor close to the origin and the input features need to be transformed. The transformation then impacts the identified PDE as it is subjected to a coordinate transformation. Using a neural network to approximate the dataset as a separate preprocessing step usually follows a pipline in which feature scaling is included, for example by preprocessing using the \texttt{Pipeline} module from \texttt{scikit-learn} \cite{scikitlearn}. It is hence important to be aware of all feature scalings in the preprocessing step and that the exact same feature scaling is used in the identification of the PDE in the second step.

Feature scaling amounts to the invertible coordinate transformations
\begin{equation}
\begin{aligned}
\tau &= \tau(t), \\
\bm{\xi} &= \bm{\xi}(\bm{x})
\end{aligned}
\label{transform}
\end{equation}
where $\tau$, $\bm{\xi} = [\xi_1, \ldots, \xi_N]$ are the new time and space coordinates, respectively. A common transformation is to shift and scale such that each input feature has zero mean and unit variance,
\begin{equation}
\begin{aligned}
\tau &= \frac{t - \bar{t}}{\sigma(t)}, \\
\bm{\xi} &= \frac{\bm{x} - \bar{\bm{x}}}{\sigma(\bm{x})}, \\
\end{aligned}
\label{shiftscale}
\end{equation}
where $\bar{t}$, $\bar{\bm{x}}$ and $\sigma(\cdot)$ denotes the (componentwise) average and standard deviation of the input data, respectively, and the division is performed componentwise where needed.

As an example we can consider what happens to the discovery of the viscous Burger's equation under the transformation \eqref{transform}. Assume we are given a dataset generated by the viscous Burger's equation in 1D,
\begin{equation}
u_t + uu_x = \epsilon u_{xx},
\label{origburger}
\end{equation}
to which we fit a neural network under the general coordinate transformation \eqref{transform}. By the chain rule we get
\begin{equation}
\begin{aligned}
\frac{\partial u}{\partial t} &= \frac{\partial u}{\partial \tau} \frac{\partial \tau}{\partial t}, \\
\frac{\partial u}{\partial x} &= \frac{\partial u}{\partial \xi} \frac{\partial \xi}{\partial x}, \\
\frac{\partial^2 u}{\partial x^2} &= \frac{\partial^2 u}{\partial \xi^2} \left(\frac{\partial \xi}{\partial x}\right)^2 + \frac{\partial u}{\partial \xi} \frac{\partial^2 \xi}{\partial x^2}
\end{aligned}
\end{equation}
and hence the neural network is not an approximation to the solution of \eqref{origburger} but to the transformed equation
\begin{equation}
\frac{\partial \tau}{\partial t} u_{\tau} + \left(\frac{\partial \xi}{\partial x} u - \epsilon \frac{\partial^2 \xi}{\partial x^2}\right)u_{\xi} = \epsilon \left(\frac{\partial \xi}{\partial x}\right)^2 u_{\xi\xi}.
\label{transformburger}
\end{equation}
Under the linear transformation \eqref{shiftscale}, the above equation reduces to
\begin{equation}
\frac{1}{\sigma(t)} u_{\tau} + \frac{1}{\sigma(x)} uu_{\xi} = \frac{\epsilon}{\sigma^2(x)}u_{\xi\xi}.
\label{shiftscaleburger}
\end{equation}
The situation becomes more complex in higher dimensions as in general we need to compute all total derivatives in the old coordinates when computing the partial derivatives in the new coordinates as
\begin{equation}
\begin{aligned}
\frac{\partial u}{\partial x_1} &= \frac{\partial u}{\partial \xi_1} \frac{\partial \xi_1}{\partial x_1} + \cdots + \frac{\partial u}{\partial \xi_N} \frac{\partial \xi_N}{\partial x_1}, \\
&\vdotswithin{=} \\
\frac{\partial u}{\partial x_N} &= \frac{\partial u}{\partial \xi_1} \frac{\partial \xi_1}{\partial x_N} + \cdots + \frac{\partial u}{\partial \xi_N} \frac{\partial \xi_N}{\partial x_N}.
\end{aligned}
\end{equation}
We write the above expression in matrix form as
\begin{equation}
\begin{bmatrix}
	\dfrac{\partial u}{\partial x_1} \\
	\vdots \\
	\dfrac{\partial u}{\partial x_N}
\end{bmatrix}
=
\begin{bmatrix}
	\dfrac{\partial \xi_1}{\partial x_1} & \cdots & \dfrac{\partial \xi_N}{\partial x_1} \\
	\vdots & \ddots & \vdots \\
	\dfrac{\partial \xi_1}{\partial x_N} & \cdots & \dfrac{\partial \xi_N}{\partial x_N}
\end{bmatrix}
\begin{bmatrix}
	\dfrac{\partial u}{\partial \xi_1} \\
	\vdots \\
	\dfrac{\partial u}{\partial \xi_N}
\end{bmatrix}
\end{equation}
where the square matrix above is the Jacobian matrix, $J$, of the coordinate transformation. Since we are interested in the PDE in the physical coordinates, we need to transform back to the original coordinates by computing the inverse of the Jacobian,
\begin{equation}
J^{-1} =
\begin{bmatrix}
	\dfrac{\partial x_1}{\partial \xi_1} & \cdots & \dfrac{\partial x_N}{\partial \xi_1} \\
	\vdots & \ddots & \vdots \\
	\dfrac{\partial x_1}{\partial \xi_N} & \cdots & \dfrac{\partial x_N}{\partial \xi_N}
\end{bmatrix}.
\end{equation}
The transformation \eqref{shiftscale} is particularly useful in high dimensions as it is linear and acts only one coordinate direction at a time, independently of the other coordinates. This means that the Jacobian is reduced to the diagonal matrix
\begin{equation}
J =
\begin{bmatrix}
	\dfrac{1}{\sigma(x_1)} & \cdots & 0 \\
	\vdots & \ddots & \vdots \\
	0 & \cdots & \dfrac{1}{\sigma(x_N)}
\end{bmatrix}
\end{equation}
and that higher-order derivatives are easily computed since each derivative of $u$ with respect to $x_i$ only yields an additional factor or $1/\sigma(x_i)$. That is, we get
\begin{equation}
\begin{aligned}
\frac{\partial u}{\partial x_i} &= \frac{1}{\sigma(x_i)} \frac{\partial u}{\partial \xi_i}, \\
\frac{\partial^2 u}{\partial x_i \partial x_j} &= \frac{1}{\sigma(x_i) \sigma(x_j)} \frac{\partial u}{\partial \xi_i \partial \xi_j} \\
&\vdotswithin{=} \\
\frac{\partial^m u}{\partial x_i \cdots \partial x_j} &= \frac{1}{\sigma(x_i) \cdots \sigma(x_j)} \frac{\partial^m u}{\partial \xi_i \cdots \partial \xi_j}
\end{aligned}
\end{equation}
for the partial derivatives up to order $m$. Transforming the partial derivatives back to the original coordinates is reduced to multiplication by a scalar which avoids the numerically unstable and computationally expensive inversion of the Jacobian matrix.

\section{Examples} \label{examples}
There are plenty of examples in previous papers which show impressive results in the accuracy of the identified PDE despite both sparse and noisy data \cite{datadriven1, datadriven2, haydensparsepde, sparsepde, neuraldiscovery1, neuraldiscovery2}. These results are all based on known PDEs on simple geometries. We will show a few examples on what happens to the identified PDE under coordinate transformations, and some potential applications in weather/climate modeling where the governing equations are unknown.

\subsection{The viscous Burger's equation in 1D} \label{burgers}
\begin{figure}[H]
\centering
\includegraphics[width=0.75\textwidth]{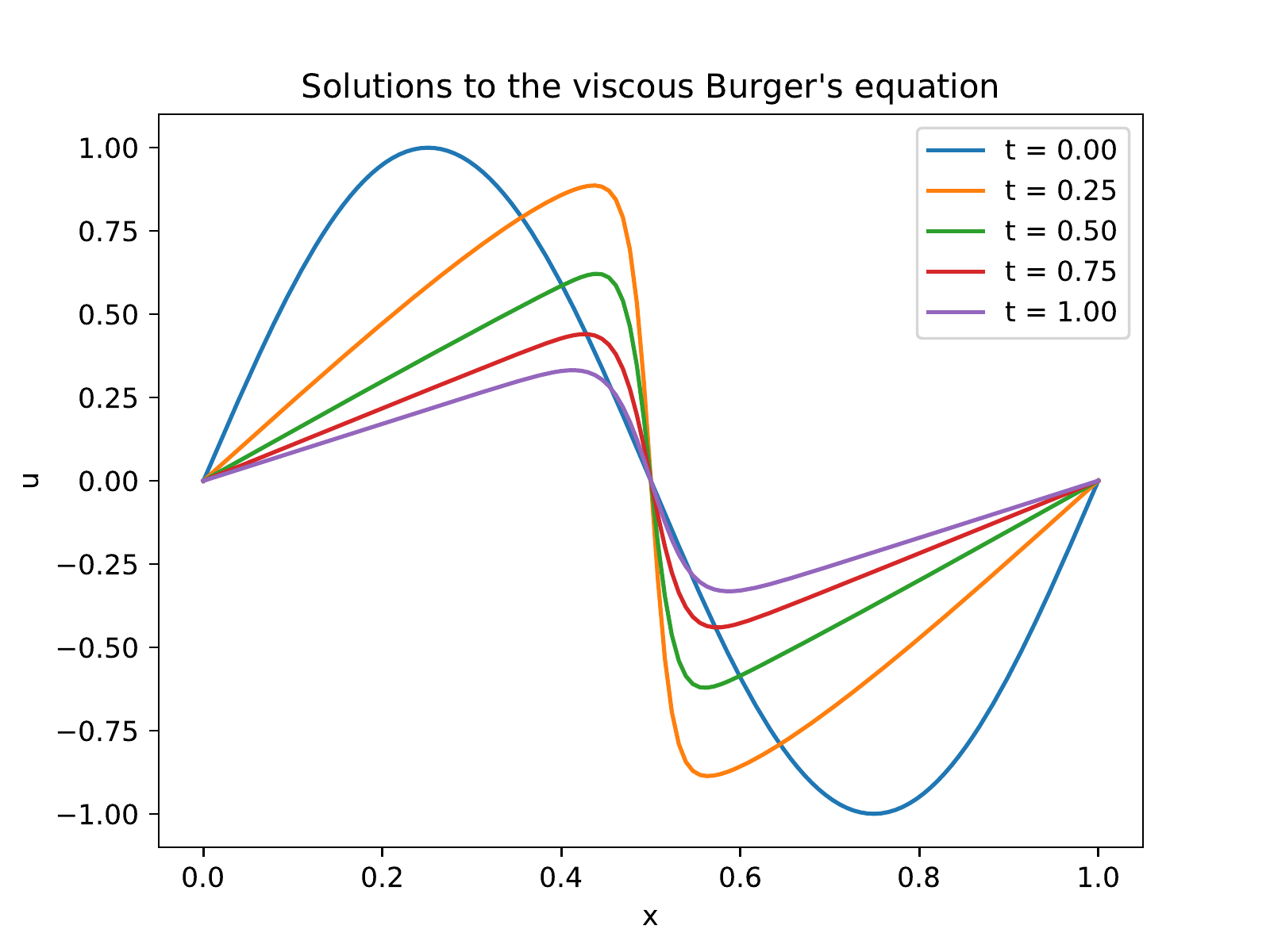}
\caption{The solution of the viscous Burger's equation forming a stationary viscous shock.}
\label{figburgersnaps}
\end{figure}

We consider the viscous Burger's equation for $(x, t)$ $\in [0, 1] \times [0, 1]$ here given by
\begin{equation}
\begin{aligned}
u_t + uu_x &= 10^{-2} u_{xx}, \\
u(0, t) &= 0, \\
u(1, t) &= 0, \\
u(x, 0) &= \sin(2 \pi x).
\end{aligned}
\label{burgerpde}%
\end{equation}
The solution to \eqref{burgerpde} is well-known and forms a decaying stationary viscous shock after a finite time, see Figure~\ref{figburgersnaps}.
The solution of \eqref{burgerpde} was computed with the finite element method using $128$ second-order elements in space and $1000$ steps using the backward Euler method in time.

To reconstruct the differential operator in \eqref{burgerpde}, we sample the solution in all interior degrees of freedom at each non-zero time step to get a dataset of the form $(t, x, u)$ consisting of a total of $255000$ entries. The first step is to fit a neural network to the dataset which allows us to compute the necessary derivatives. This is a separate preprocessing step in which we use a feedforward neural network with 5 hidden layers and 10 neurons in each layer with the hyperbolic tangent activation function. The network is trained using the BFGS method from \texttt{SciPy}'s \texttt{scipy.minimize} module with default parameters \cite{scipy}. For this model problem we consider three different parametrizations of $\hat{L}$ without regularization or scaling: 1) A linear model with the library terms $uu_x$ and $u_{xx}$ as input features, 2) A linear model with up to second order derivative and non-linear terms as input features, and 3) A two layer feedforward neural network with 2 neurons in each layer with the hyperbolic tangent activation and up to second order derivative terms as input features. When minimizing \eqref{minpde} with the different parametrizations we discover the following PDEs:
\begin{equation}
\begin{aligned}
1) && u_t &= -9.9792\times 10^{-1}*uu_{x} + 9.9596\times 10^{-3}*u_{xx} \\
\\
2) && u_t &= -9.9718\times 10^{-1}*uu_{x} + 1.0134\times 10^{-2}*u_{xx} \\
&&&- 4.2757\times 10^{-8}*\left(u_{xx}\right)^2 + 1.0156\times 10^{-5}*u_{x}u_{xx} \\
&&&- 8.3758\times 10^{-5}*uu_{xx} + 2.8494\times 10^{-6}*\left(u_{x}\right)^2 \\
&&&+ 1.4114\times 10^{-4}*u_{x} - 4.4878\times 10^{-3}*\left(u\right)^2 + 2.2429\times 10^{-3}*u \\
\\
3) && u_t &= -2.2618*\tanh(-1.0809*\tanh(5.2229\times 10^{-3}*u_{xx} \\
&&&+ 3.4171\times 10^{-1}*u_{x} + 7.0587\times 10^{-2}*t - 1.4968\times 10^{-1}*u \\
&&&- 6.2141\times 10^{-2}*x - 4.9887\times 10^{-1}) - 1.5738\times 10^{1}*\tanh(5.9850\times 10^{-3}*u_{xx} \\
&&&+ 1.1323\times 10^{-2}*u_{x} + 5.1961\times 10^{-1}*t - 3.6736\times 10^{-1}*u \\
&&&+ 6.7682\times 10^{-2}*x + 9.3685\times 10^{-1}) + 1.4128\times 10^{1}) \\
&&&- 1.4500*\tanh(-1.0331*\tanh(5.2229\times 10^{-3}*u_{xx} + 3.4171\times 10^{-1}*u_{x} \\
&&&+ 7.0587\times 10^{-2}*t - 1.4968\times 10^{-1}*u - 6.2141\times 10^{-2}*x - 4.9887\times 10^{-1}) \\
&&&+ 1.9307\times 10^{1}*\tanh(5.9850\times 10^{-3}*u_{xx} + 1.1323\times 10^{-2}*u_{x} \\
&&&+ 5.1961\times 10^{-1}*t - 3.6736\times 10^{-1}*u + 6.7682\times 10^{-2}*x + 9.3685\times 10^{-1}) \\
&&&- 1.7623\times 10^{1}) + 1.7249\times 10^{-1}
\end{aligned}
\end{equation}
It is clear that the different models have different trade-offs. The first model is similar in apperance to the true PDE, but it is required that we know the form of the PDE a priori. The second model has small coefficients for the spurious terms and close to the true values for the true terms. The third model is general and of limited use for human interpretation. However, many PDE solvers offer automatic discretization of symbolic expressions and the output of the general model can be used as input to a software such as Comsol Multiphysics \cite{comsol} or physics informed neural networks \cite{2017arXiv171110561R}.

To see the effect of a feature scaling we consider the simple library model under the standard shift and scale transformation \eqref{shiftscale}. For this particular dataset we have
\begin{equation}
\begin{aligned}
\sigma^2(t) &= 0.0833326, & \sigma^2(x) &= 0.08268167, \\
\bar{t} &= 0.50050196, & \bar{x} &= 0.49999807,
\end{aligned}
\end{equation}
and the identified PDE in transformed space becomes
\begin{equation}
u_{\tau} = -1.0010uu_{\xi} + 3.4815\times 10^{-2}u_{\xi\xi}
\end{equation}
which corroborates \eqref{shiftscaleburger} rewritten as
\begin{equation}
u_{\tau} = -\frac{\sigma(t)}{\sigma(x)}uu_{\xi} + \frac{\sigma(t)}{\sigma^2(x)} \times 10^{-2} u_{\xi\xi}.
\end{equation}
To get the PDE in the physical coordinates it is hence required that we invert the coordinate transformation and compute the derivatives in the physical space as
\begin{equation}
\begin{aligned}
\frac{\partial u}{\partial \tau} &= \frac{\partial u}{\partial t} \frac{\partial t}{\partial \tau} = \sigma(t) \frac{\partial u}{\partial t} \\
\frac{\partial u}{\partial \xi} &= \frac{\partial u}{\partial x} \frac{\partial x}{\partial \xi} = \sigma(x)\frac{\partial u}{\partial x} \\
\frac{\partial^2 u}{\partial \xi^2} &= \frac{\partial^2 u}{\partial x^2} \left(\frac{\partial x}{\partial \xi}\right)^2 = \sigma^2(x)\frac{\partial^2 u}{\partial x^2}.
\end{aligned}
\end{equation}
First after transforming back to the physical space do we recover the desired PDE
\begin{equation}
u_t + \frac{\sigma(x)}{\sigma(t)}uu_x = \frac{\sigma^2(x)}{\sigma(t)} u_{xx},
\end{equation}
and in this particular case we get
\begin{equation}
u_t + 0.99708uu_x = 0.99717 \times 10^{-2}u_{xx}.
\end{equation}
For this model problem, coordinate transformations are not necessary as we are working on the simple domain $(x,t) \in [0, 1] \times [0, 1]$ which is in the range where machine learning algorithms performs well.

\subsection{Temperature distribution in 2D} \label{temperature}
The focus in this section is on potential applications of the method to real measurement data for weather/climate modeling. The outlined method should be seen as a starting point for further research.

\begin{figure}[H]
\centering
\begin{subfigure}[t]{0.49\textwidth}
\centering
\includegraphics[scale=0.4]{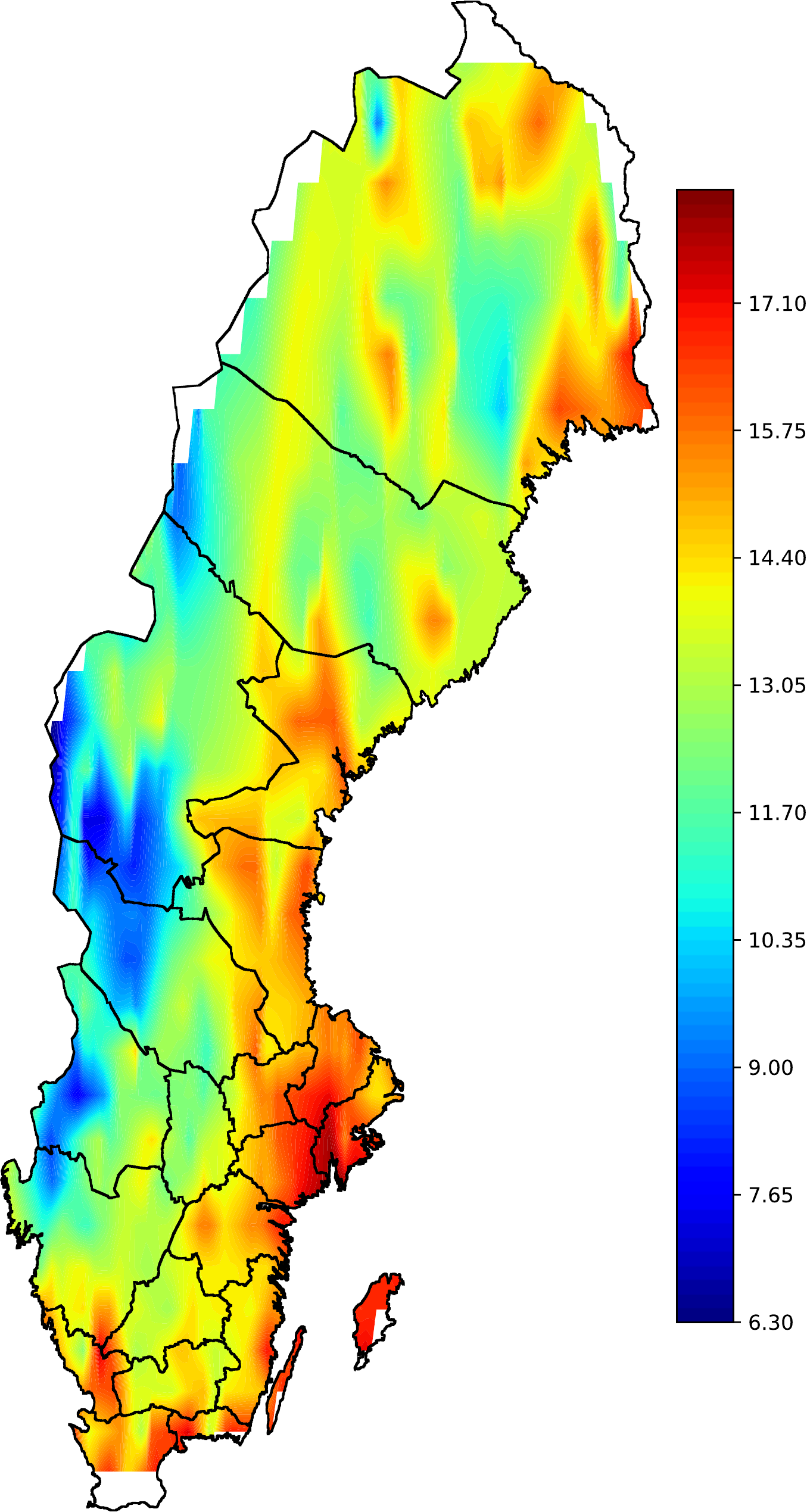}
\caption{Linear interpolation temperature snapshot.}
\end{subfigure}
\begin{subfigure}[t]{0.49\textwidth}
\centering
\includegraphics[scale=0.4]{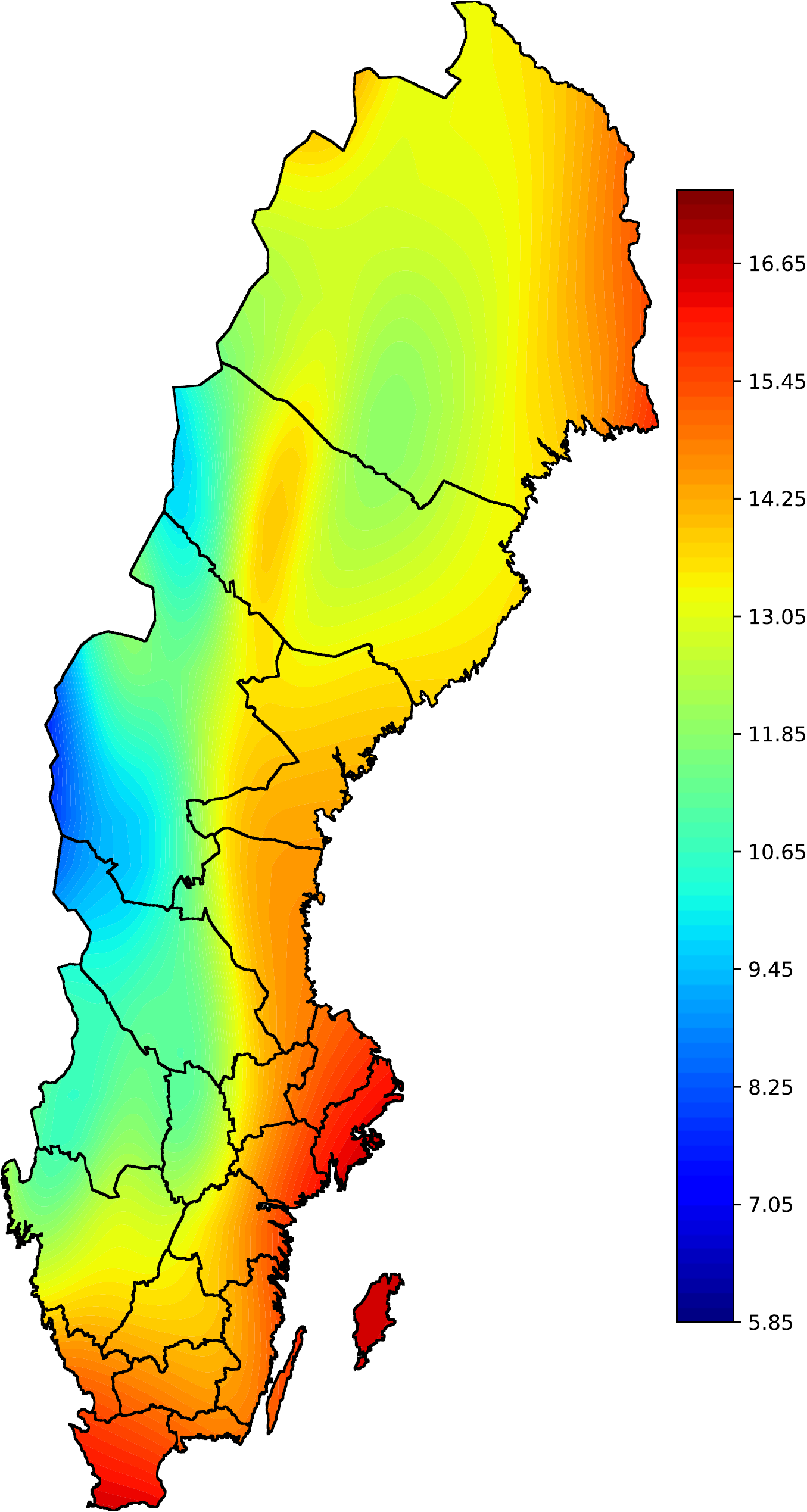}
\caption{Neural network approximated temperature snapshot.}
\end{subfigure}
\caption{The linear interpolation and neural network approximated temperature snapshots. The neural network has 5 layers with 20 neurons each.}
\label{figsmhistations}
\end{figure}

A natural application of the method is where several geostationary sensors are recording measurements over time, for example weather stations which measure quantities such as temperature, pressure, humidity, and so on on a regular basis. The Swedish Meteorological and Hydrological Institute\footnote{\url{http://www.smhi.se}} is offering a REST API where meteorological data can be downloaded for all 326 measurement stations in Sweden. Each station is recording data at time intervals ranging from every hour to every 12 hours, and the locations are given in latitude/longitude coordinates in the range $[10.96, 55.34] \times [24.17, 69.05]$ which is outside the range where machine learning algorithms perform well. We downloaded the data and made a dataset consisting of the temperature for the first week in July 2016 to see if we can find a PDE which is describing the temperature distribution. The dataset contains irregular measurements in a complicated geometry where coordinate transformations are inevitable. The dataset is imbalanced since there are too many points in time compared to the number of points in space. In this artificial example, we remedy this by performing a linear interpolation in space and time onto the convex hull of a regular grid with 168 time points, 32 latitude points, and 128 longitude points, see Figure~\ref{figsmhistations} (where all spatial data points have been transformed by the Mercator projection for visualization only). The final interpolated dataset contains 688129 data points on a regular grid. Finally, we approximate the dataset with a neural network with 5 hidden layers with 20 neurons in each layer using the L-BFGS optimization method. We tried many different networks and this, surprisingly small network, had the best generalization accuracy when evaluated on different test sets obtained by different interpolations. Larger networks had problems with overfitting and adding dropout and regularization caused the L-BFGS algorithm to perform poorly.
\begin{figure}[H]
\centering
\begin{subfigure}[t]{0.49\textwidth}
\centering
\includegraphics[scale=0.45]{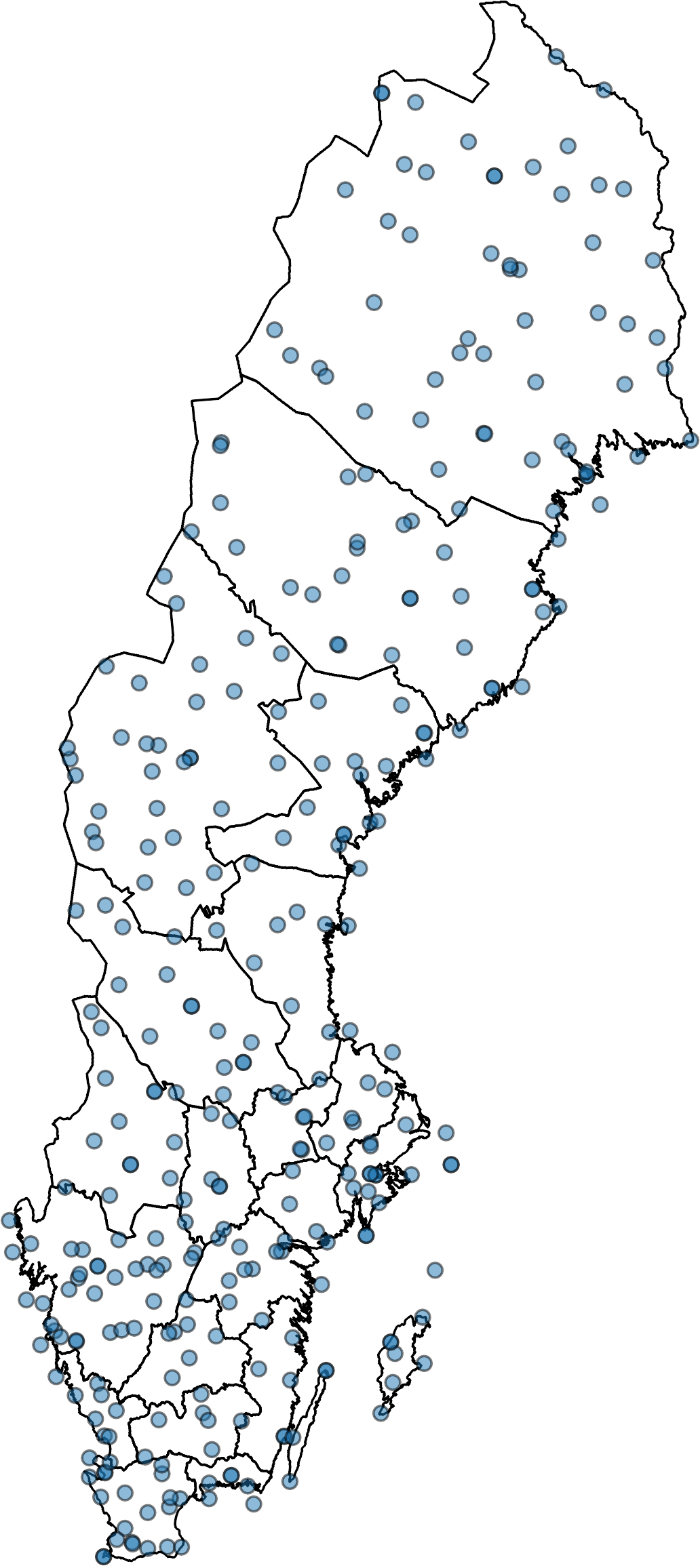}
\caption{The locations of the 326 weather stations in Sweden.}
\end{subfigure}
\begin{subfigure}[t]{0.49\textwidth}
\centering
\includegraphics[scale=0.45]{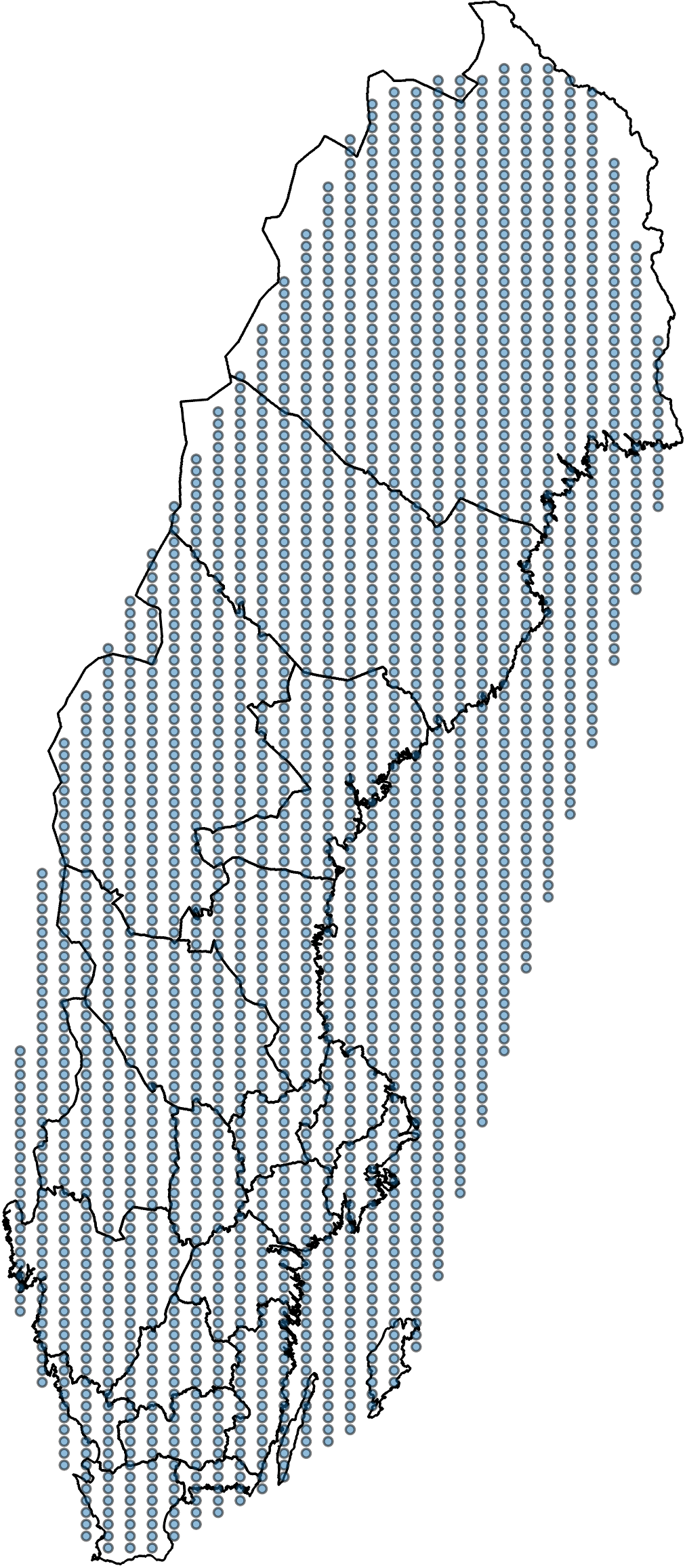}
\caption{Interpolation onto the convex hull of the weather station locations.}
\end{subfigure}
\caption{Physical and interpolated locations of the geostationary locations.}
\end{figure}

Note that since the neural network is globally defined we can plot the temperature in the whole domain and not just on the convex hull of the data points. In this case, the governing PDE is unknown and we will elaborate on results and conclusions in section~\ref{model}.
\begin{remark}
The linear interpolation of the dataset of course gives rise to non-physical linear relations in the neural network approximation. In a real case one should consider more advanced reconstruction methods if more data is needed.
\end{remark}

\section{Feature selection} \label{feature}
To elaborate on feature selection we return to Section~\ref{burgers} and the polynomial PDE model for the viscous Burger's equation which has a decent trade-off between complexity and interpretability. By adding L1 regularization to the polynomial PDE model with $\alpha_q = 10^{-2}$ in \eqref{minpde}, the spurious terms are further reduced to
\begin{equation}
\begin{aligned}
u_t &=  -9.9473\times 10^{-1}*uu_{x} + 1.0105\times 10^{-2}*u_{xx} \\
&- 1.2630\times 10^{-9}*\left(u_{xx}\right)^2 + 1.0007\times 10^{-5}*u_{x}u_{xx} \\
&- 5.2975\times 10^{-5}*uu_{xx} - 3.3428\times 10^{-5}*\left(u_{x}\right)^2 \\
&+ 1.2649\times 10^{-6}*u_{x} - 1.5698\times 10^{-5}*\left(u\right)^2 - 1.6640\times 10^{-6}*u
\end{aligned}
\end{equation}
which can be removed by some predefined cut-off value for the coefficient size.
\begin{remark}
As the polynomial PDE model is linear we can, of course, use the traditional least squares method with Lasso \cite{lasso} instead of adding L1 regularization to the optimization problem. In that case we obtain the even sparser model
\begin{equation}
\begin{aligned}
u_t &=  -9.9216\times 10^{-1}*uu_{x} + 1.0082\times 10^{-2}*u_{xx} \\
&- 2.3627\times 10^{-9}*\left(u_{xx}\right)^2 + 1.0129\times 10^{-5}*u_{x}u_{xx} \\
&- 5.4086e\times 10^{-5}*uu_{xx} - 3.2458\times 10^{-5}*\left(u_{x}\right)^2.
\end{aligned}
\end{equation}
The traditional least squares model does not, however, generalize to differential operators of arbitrary complexity or very large datasets.
\end{remark}

It is common to analyze the input data in order to remove redundant or correlated features. In this case, it is only the terms $u$, $u_{x}$, and $u_{xx}$ which are independent. A common method is to compute the variance of the input data and remove features with low variance since they are deemed as unimportant. This method does not apply in a PDE context since high order derivatives have lower regularity and hence usually a higher variance, which is clearly shown in Table~\ref{tablefeatures}. More sophisticated methods for feature selection include stability analysis via randomized Lasso (R-Lasso) \cite{rlasso}, recursive feature elimination (RFE) \cite{rfe}, and Boruta \cite{boruta}. We include comparisons with the two former methods in Table~\ref{tablefeatures} were we have used the implementations from \texttt{scikit-learn} with default parameters. The Boruta method works on ensamble models, such as random forests, and is not suitable in this context. We did, however, try the Boruta method on our dataset with a random forest regressor and we did not obtain any good results. The Boruta method deemed all features as equally important.
\def\arraystretch{1.5}
\begin{table}[H]
\centering
\begin{tabular}{|c|c|c|c|c|c|c|c|c|c|}
\hline
Feature & $u$ & $u_{x}$ & $u_{xx}$ & $u^2$ & $uu_{x}$ & $uu_{xx}$ & $u^2_{x}$ & $u_{x}u_{xx}$ & $u^2_{xx}$ \\
\hline
Variance & $0.21$ & $23$ & $23000$ & $0.06$ & $3.1$ & $5700$ & $11000$ & $9.0 \times 10^{6}$ & $1.7 \times 10^{10}$ \\
R-Lasso & $0.09$ & $0$ & $1$ & $1$ & $1$ & $1$ & $1$ & $1$ & $0$ \\
RFE & $3$ & $5$ & $2$ & $4$ & $1$ & $6$ & $7$ & $8$ & $9$ \\
\hline
\end{tabular}
\caption{The variance, feature importance and feature ranking of our dataset for the viscous Burger's equation.}
\label{tablefeatures}
\end{table}
\def\arraystretch{1.0}
We can see from Table~\ref{tablefeatures} that the variance of the features are the opposite of what is expected as the variance grows with the order of the derivative independent of the importance of the feature. By combining R-Lasso and RFE we can get a decent understanding of which features that are important in the dataset.

\section{Model selection} \label{model}
\begin{figure}[H]
\centering
\includegraphics[width=\textwidth]{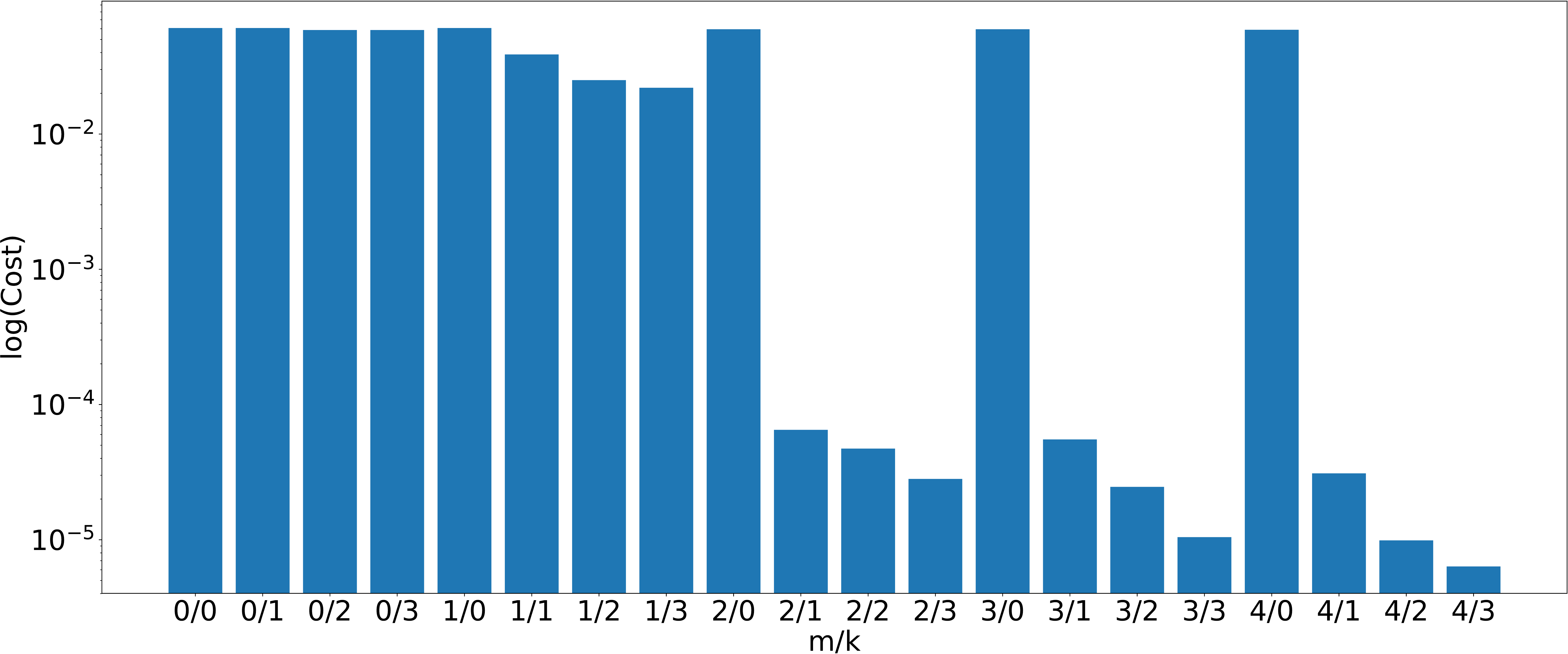}
\caption{The logarithm of the cost function for different choices of derivative and non-linear orders $m$ and $k$ for the viscous Burger's equation. The true configuration is $m/k = 2/1$.}
\label{figburgerbars}
\end{figure}

\begin{figure}[H]
\centering
\includegraphics[width=\textwidth]{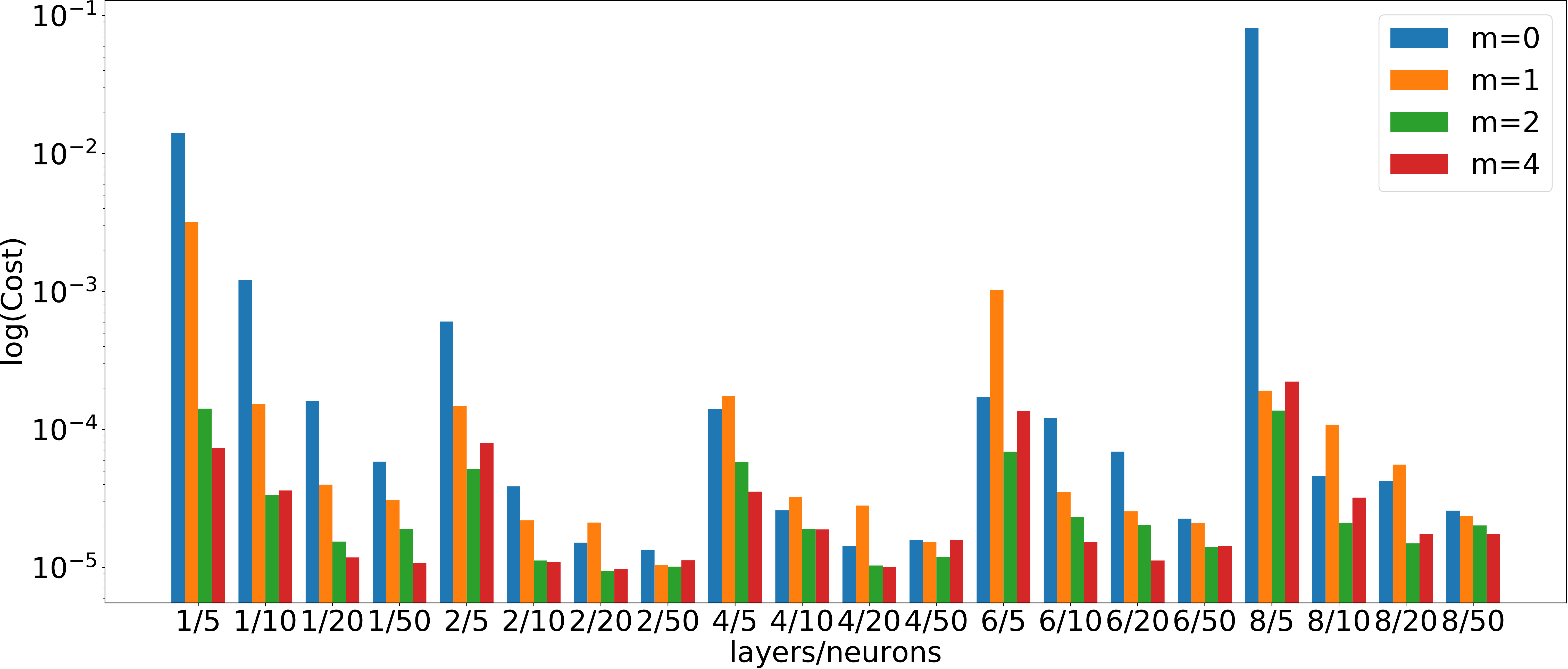}
\caption{The logarithm of the cost function for different network designs and partial derivative orders ($m$) for the viscous Burger's equation.}
\label{figburgerbarsnetwork}
\end{figure}

\begin{figure}[H]
\centering
\begin{subfigure}[t]{0.49\textwidth}
\centering
\includegraphics[width=\textwidth]{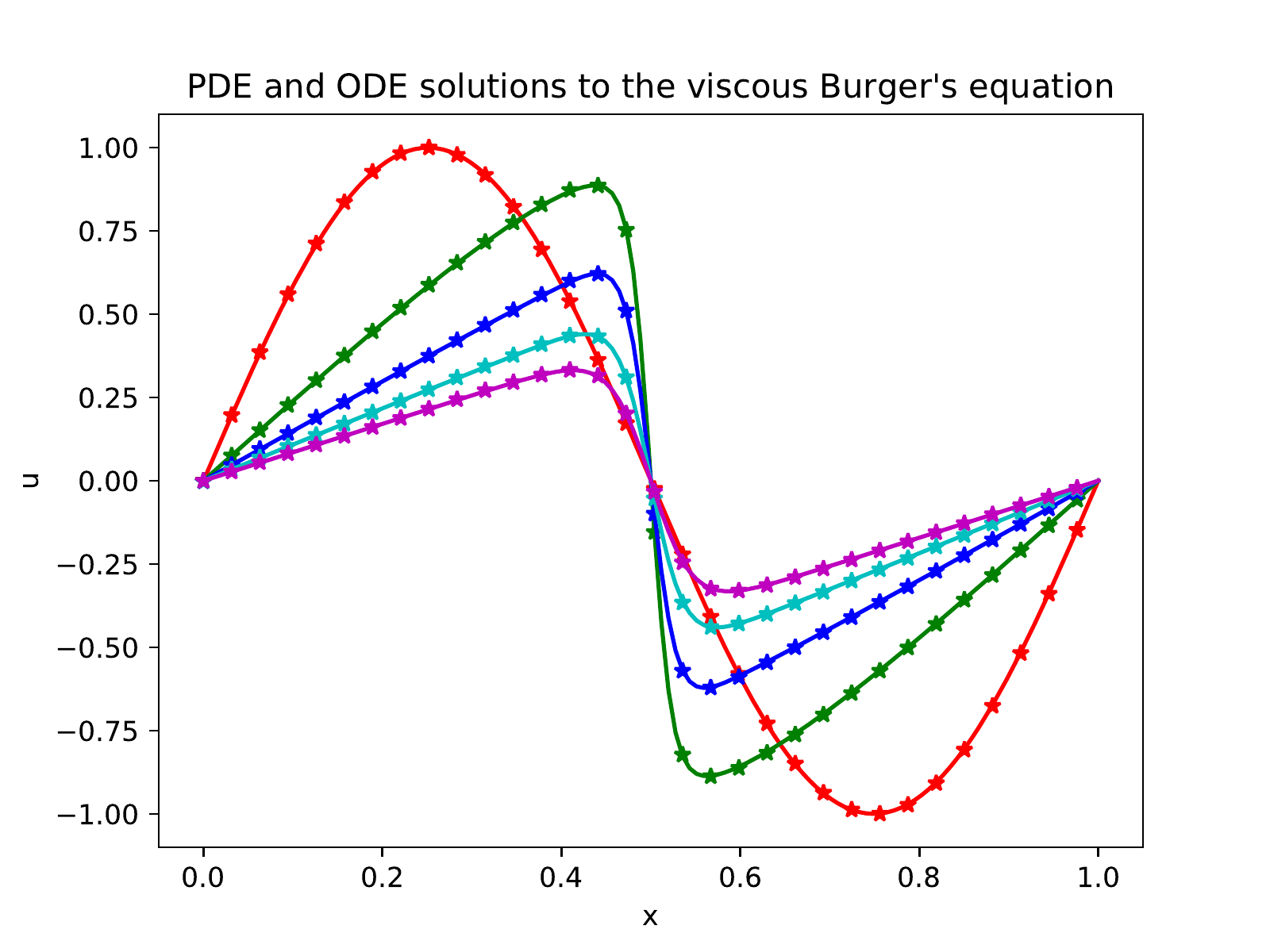}
\caption{Line: PDE solution, Dots: ODE solution. The PDE and ODE solutions to the viscous Burger's equation at times 0.0, 0.25, 0.5, 1.0.}
\end{subfigure}
\begin{subfigure}[t]{0.49\textwidth}
\centering
\includegraphics[width=\textwidth]{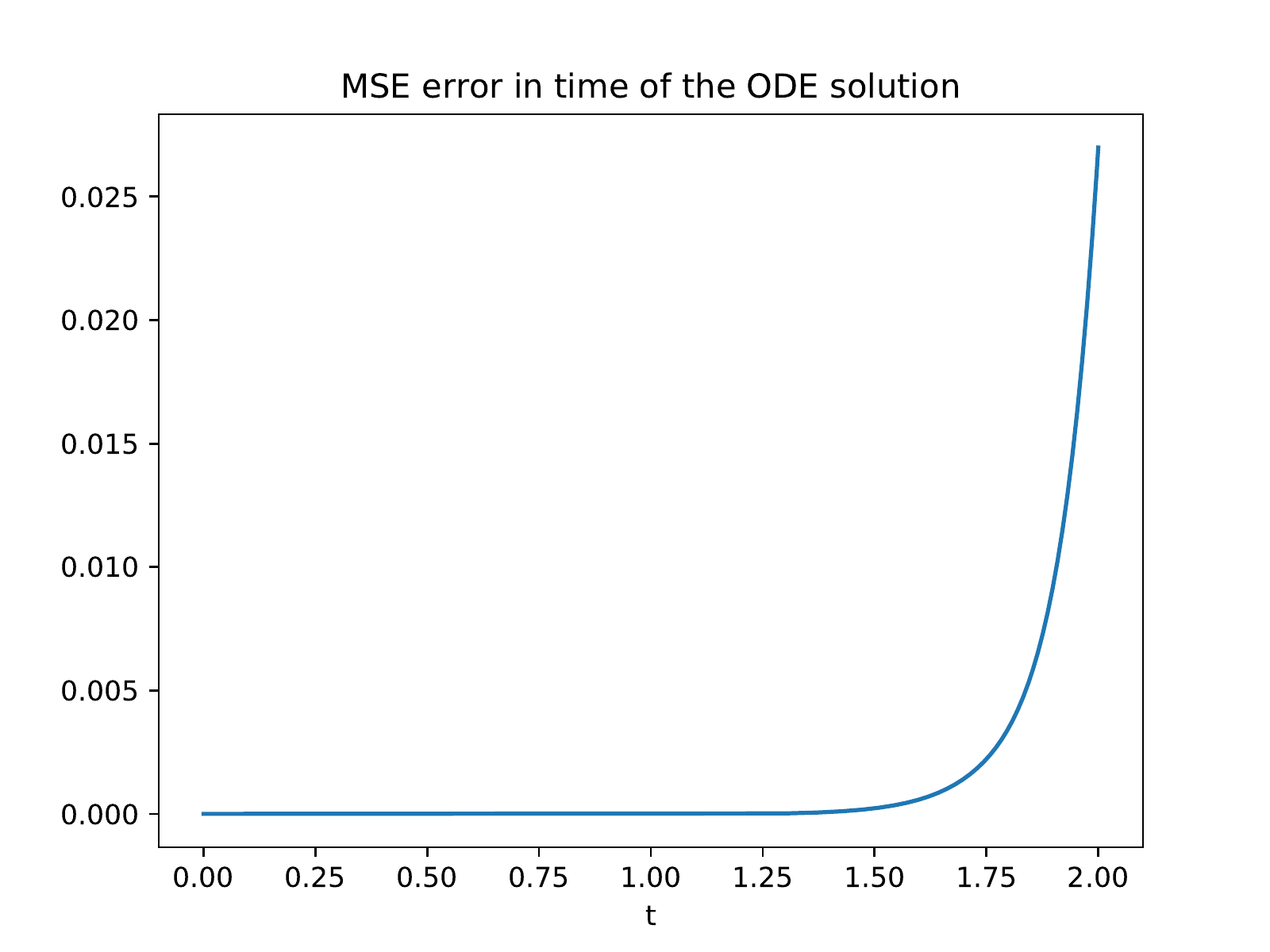}
\caption{The mean square error in time of the ODE solution to the viscous Burger's equation.}
\end{subfigure}
\caption{Comparison between the ODE and PDE solutions of the viscous Burger's equation. The ODE solution is accurate for $0 \leq t \leq 1$ where we have trained the operator. The ODE operator is, however, unable to extrapolate for $t >> 1$.}
\label{figburgerode}
\end{figure}

As the polynomial PDE model for the viscous Burger's equation is linear and the optimization problem \eqref{minpde} is convex, minimization using standard least squares or gradient based optimization is efficient and model selection can be performed by an exhaustive parameter search. By computing the value of the cost function for different choices of the derivative order $m$ and non-linear order $k$, it is clearly seen when a suitable model has been found. In Figure~\ref{figburgerbars} we show the logarithm of the cost function for different choices of $m$ and $k$. We can see that the cost function is instantly reduced by several orders of magnitude when a sufficient model has been found.
We can perform a similar study when the PDE is represented by a neural network with different number of layers and neurons. In Figure~\ref{figburgerbarsnetwork} we show the value of the cost function for different network designs with different partial derivative orders as input.
The case with 2 hidden layers with 50 neurons in each layer is particularly interesting. In this case we have a low cost even without any partial derivatives as input. Thus for the case $m=0$, the viscous Burger's equation is effectively transformed into an ordinary differential equation (ODE) of the form
\begin{equation}
\hat{u}_t = \hat{L}(\hat{u}).
\label{ode}
\end{equation}
The ODE \eqref{ode} can easily be solved using any time integration method. In Figure~\ref{figburgerode} we used standard Runge-Kutta 4(5) from \texttt{SciPy} with default settings to integrate the ODE. We can see that the ODE operator gives accurate results for $0 \leq t \leq 1$ where we have trained the operator. We can also see, unfortunately, that the ODE operator is unable to extrapolate far beyond $t=1$ where we have no training data. It is, however, quite remarkable that the dynamics of a second order non-linear PDE can be well approximated by an ODE in the range of the training data and slightly beyond.

\begin{figure}[H]
\centering
\includegraphics[width=\textwidth]{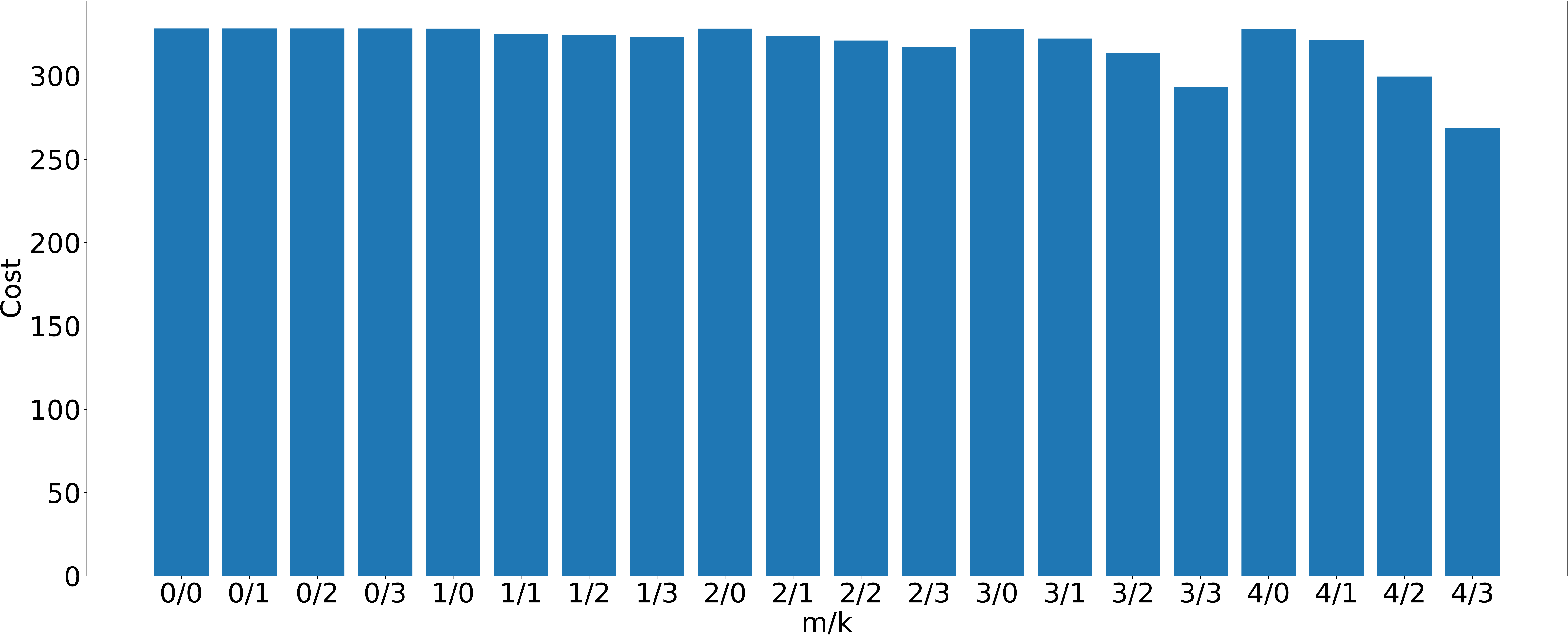}
\caption{The value of the cost function for different choices of derivative and non-linear orders $m$ and $k$ for temperature models. No sufficient polynomial models were found.}
\label{figsmhibarsclassic}
\end{figure}

\begin{figure}[H]
\centering
\includegraphics[width=\textwidth]{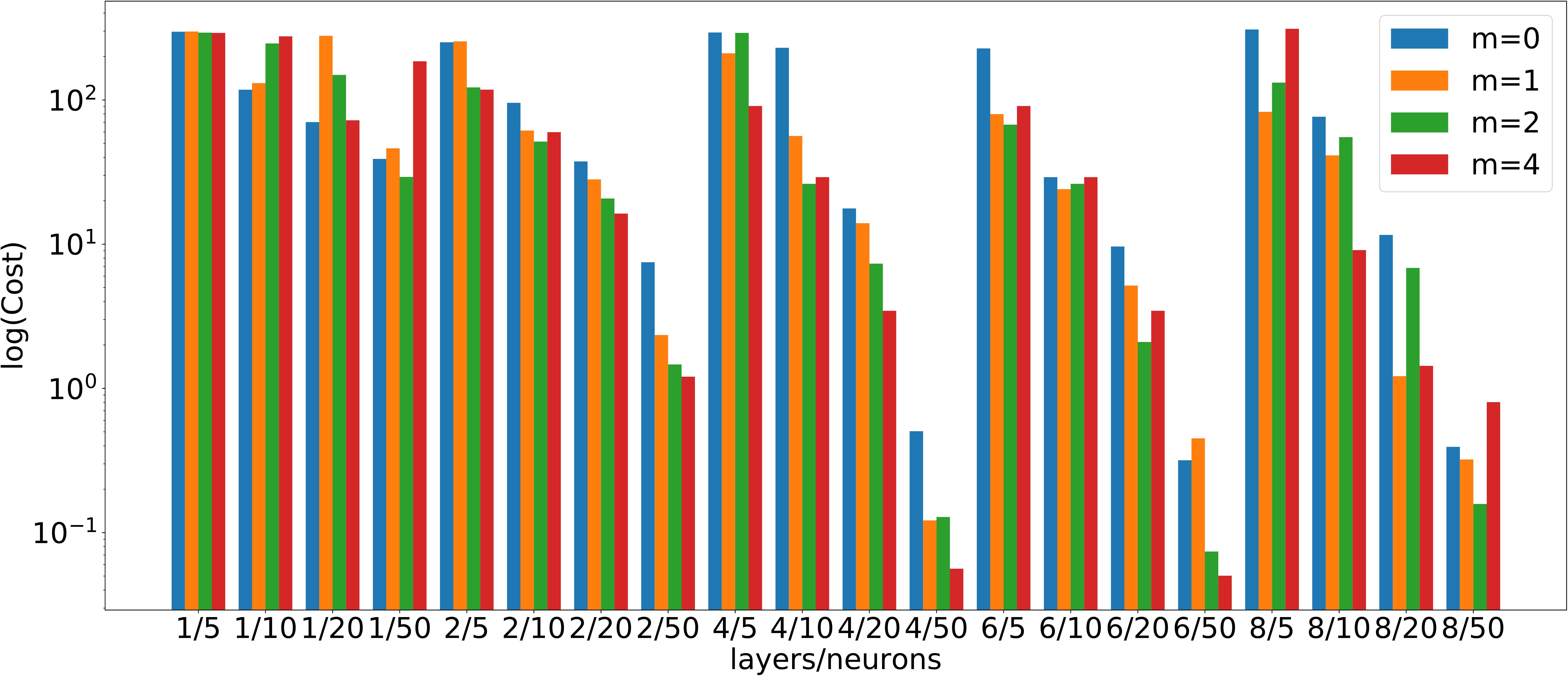}
\caption{The logarithm of the cost function for different network architectures and partial derivative orders ($m$) for temperature models. Some sufficient network models were found.}
\label{figsmhibarsnetwork}
\end{figure}

\begin{figure}[H]
\centering
\includegraphics[width=0.75\textwidth]{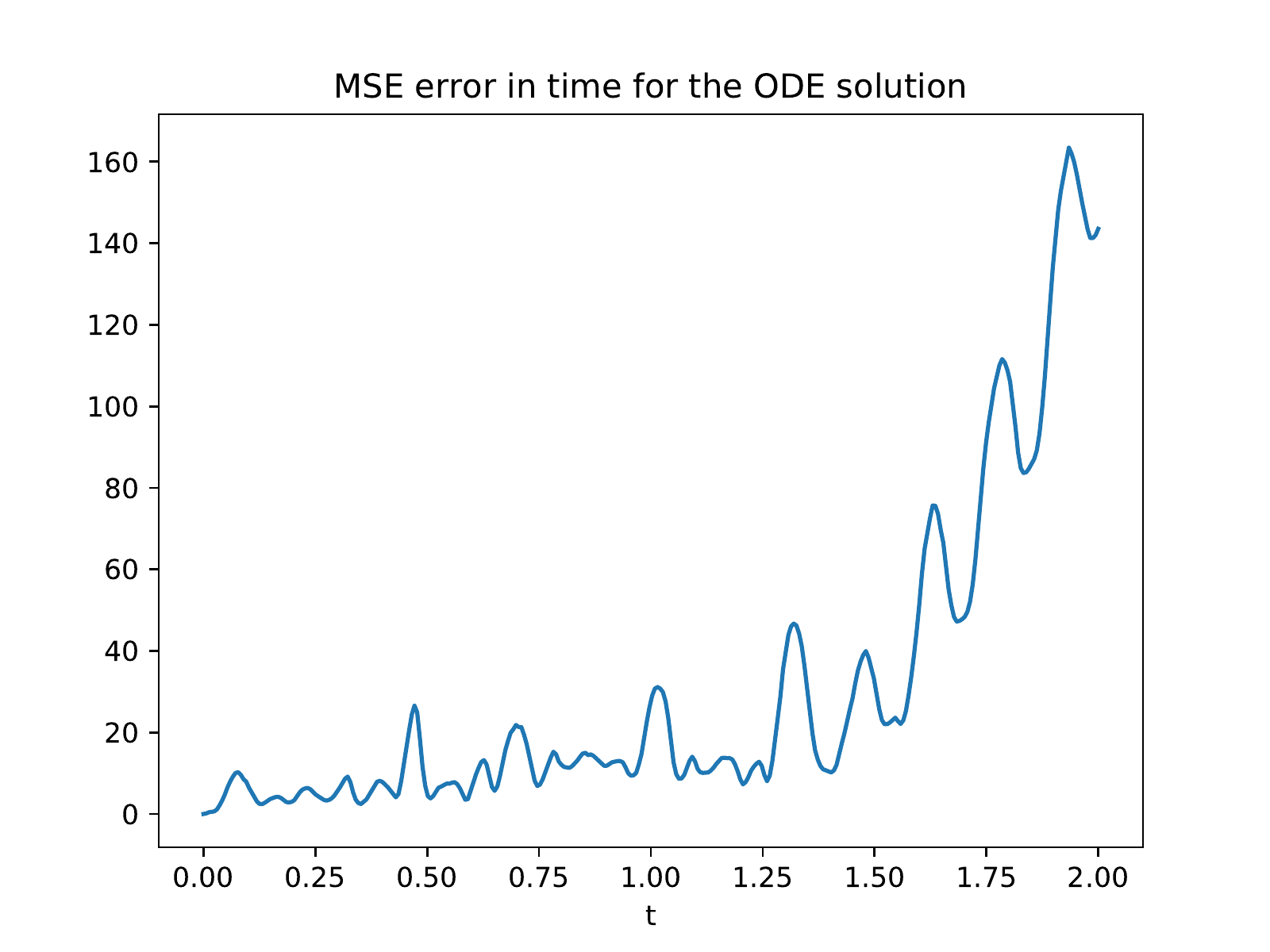}
\caption{The mean square error in time of the ODE temperature model.}
\label{figsmhiode}
\end{figure}

This method can in the same way be used for model invalidation. Since a PDE model for the temperature distribution is unknown we can perform an exhaustive parameter search to see if a sufficient model can be found. In Figure~\ref{figsmhibarsclassic} we show the value of the cost function for different values of $m$ and $k$, and we can clearly see that there is no sufficient model in this parameter range.
Since no polynomial PDE models for the temperature distribution were found, we can perform the same exhaustive parameter search where we instead vary the number of layers and neurons in each layers when $\hat{L}$ is represented by a neural network. The results can be seen in Figure~\ref{figsmhibarsnetwork} where we represented $\hat{L}$ by neural networks with 1, 2, 4, 6, 8 hidden layers with 5, 10, 20, 50 neurons in each layer, respectively, and partial derivatives of order $m=0, 1, 2, 4$ as input. We can see that the cost drops several orders of magnitude for certain configurations which indicate that sufficient models have been found. We can also see that even in this complicated case, there are some ODE models which appears to capture the dynamics.
Similarly to viscous Burger's case, we use the ODE operator with 6 layers and 50 neurons in each layer to compute the mean square error in time for the ODE solution using Runge-Kutta 4(5). In this case, the ODE operator is trained on data from the first week in July 2016 ($0 \leq t \leq 1$) and evaluated on both the first and second week ($0 \leq t \leq 2$) to test the prediction performance. As in the viscous Burger's case, we can see in Figure~\ref{figsmhiode} that the ODE operator is fairly accurate in the region where training data is available but is unable to extrapolate far beyond the training data. However, the operator is able to remain accurate up to time $t=1.25$ which amounts to quarter of a week in physical time.
The simulation shown in Figure~\ref{figsmhiode} of the Swedish temperature distribution over a two week period using the ODE operator takes only a fraction of a second on a laptop. We hence believe that by incorporating more quantities in the measurements, it is possible to discover a system of ODEs which can be used to obtain both fast and accurate short-time predictions.

\section{Summary and conclusions} \label{summary}
We have used deep artificial neural networks to discover partial differential equations from data sets consisting of measurements of physical quantities of interest. The quantities of interest are both artificial from known model PDEs, as well as true measurement data from weather stations.

In general, the physical domain is non-trivial and data transformations are necessary to bring the problem into a range where machine learning algorithms perform well. These data transformations amounts to coordinate transformations in the discovered PDEs and it is hence important that all data transformations are recorded such that the discovered PDEs can be transformed back into physical space. We have shown examples of general data transformations and the common shift and scale transformation in particular.

The discovered PDE operator is not unique for any given data set. We performed parameter searches to discover a range of operators that describes a PDE which is generating our data set. We found that the dynamics of the non-linear, second order viscous Burger's equation could also be well approximated by an ODE which was automatically discovered. We also found an ODE for a 2D temperature distribution model which shows interesting properties for further research. The ODE operators we found are accurate in the region of the training data and are able to extrapolate slightly beyond the training data. The benefit of the ODE models is that they can be solved in fractions of a second on a laptop, compared to the PDE models which require substantial computational resources.

\section{Acknowledgements} \label{acknowledgements}
Some of the computations were performed on resources provided by The Swedish National Infrastructure for Computing (SNIC) through Uppsala Multidisciplinary Center for Advanced Computational Science (UPPMAX) under Project SNIC 2017/7-131.

The authors were partially supported by a grant from the G{\"o}ran Gustafsson Foundation for Research in Natural Sciences and Medicine.

\bibliographystyle{abbrv}
\bibliography{citings}

\end{document}